\documentclass[preprint,12pt]{elsarticle}

\usepackage{amssymb}
\usepackage{amsmath}
\usepackage{xcolor}
\usepackage{booktabs}

\journal{Nuclear Physics B}

\begin{document}

\begin{frontmatter}

\title{Toward Reliable Infant Pose Estimation: A Training-Dynamics Approach to Noisy Annotation Detection}

\author[1]{Emanuele Cardinale\corref{cor1}}
\ead{emanuele.cardinale@phd.unich.it}
\cortext[cor1]{Corresponding author}
\author[2]{Sara Moccia}
\author[3]{Alessandro Cacciatore}
\author[4]{Lucia Migliorelli}

\affiliation[1]{organization={Department of Engineering and Geology, Università degli Studi ``G. d’Annunzio'' Chieti-Pescara},
                city={Pescara},
                postcode={65127},
                country={Italy}}

\affiliation[2]{organization={Laboratory of Computational Oncology, Department of Oncology, KU Leuven},
                city={Leuven},
                postcode={3000},
                country={Belgium}}

\affiliation[3]{organization={Department of Political Science, Università di Teramo},
                city={Teramo},
                postcode={64100},
                country={Italy}}

\affiliation[4]{organization={Department of Innovative Technologies in Medicine and Dentistry, Università degli Studi ``G. d’Annunzio'' Chieti-Pescara},
                city={Chieti},
                postcode={66100},
                country={Italy}}



\begin{abstract}
Spontaneous movement analysis in preterm infants relies increasingly on markerless pose estimation (PE) to derive clinically relevant motion biomarkers directly from video recordings. Training accurate infant PE models requires large sets of manually annotated keypoints, and human annotation is inherently prone to error. Noisy keypoints (i.e., keypoints mislocalized with respect to their true anatomical position) are especially problematic in this clinical setting, since they can propagate as artificial artifacts into the reconstructed joint trajectories. Building on the small-loss hypothesis and training-dynamics-based sample selection established in the noisy-label learning literature, we propose a novel framework for detecting noisy keypoint annotations. A hybrid convolutional-attention model is trained to predict the anatomical category of each keypoint from its spatial coordinates and local visual features; the resulting cross-entropy training dynamics are then used to derive per-keypoint descriptors, which are partitioned into clean and noisy subsets via unsupervised clustering. We validate the approach on NeoPose, a newly collected dataset of 65 hospitalized preterm infants, under two realistic noise scenarios (random positional perturbation and left–right swapping) across multiple noise levels. Results show that the proposed approach achieves an F1-score of up to 91.9\% in noisy-keypoint detection. The framework further generalizes to the heterogeneous COCO benchmark, where filtering CE-detected noisy keypoints from the training set also yields measurable improvements (up to 7.4 AP points) in downstream pose estimation accuracy at moderate-to-high noise levels.
\end{abstract}





\end{frontmatter}



\section{Introduction}

Spontaneous, endogenously-generated movement patterns exhibited by preterm infants (i.e., newborns born before 37 weeks of gestation) during post-birth hospitalization are highly reliable markers for the early assessment of neurodevelopmental trajectories \cite{gm_neurosviluppo1,gm_neurosviluppo2}. During the first months of life, infants display a rich repertoire of spontaneous movements that reflect the integrity of the nervous system, before intentional and purposeful movements begin to emerge. Abnormalities in the quality, complexity, or variability of these movements may indicate early neurological dysfunction and are associated with a range of conditions, such as cerebral palsy and autism spectrum disorder \cite{gma_autism1,gma_autism2,gma_autism3}. Early identification of atypical motor patterns is particularly critical for preterm infants, who are at significantly higher risk of neurodevelopmental impairment and long-term disability \cite{preterm_gma}. The clinical assessment of infants' spontaneous movements currently relies on the visual inspection of trained experts, who evaluate motor patterns through direct observation \cite{gma_autism4}. However, despite its strong predictive value, this approach is inherently subjective, time-consuming, and dependent on the individual clinician's expertise, making it prone to inter-observer variability and limiting its scalability in routine clinical practice.

In recent years, computer-assisted approaches based on markerless pose estimation (PE) and quantitative motion analysis from conventional 2D RGB video recordings have emerged as unobtrusive tools to support clinicians in this assessment. These approaches leverage PE outputs to derive quantitative motion descriptors from the reconstructed skeletal representation of the infant's body, including joint angle trajectories, limb velocity profiles, movement complexity and variability measures, and symmetry indices between homologous body segments \cite{gma_1,gma_2,gma_3}. Current state-of-the-art PE models are trained on adult-centric datasets and therefore require adaptation to account for the unique anatomical proportions of infants and the diverse motor repertoires exhibited while lying in the supine position (i.e., the posture most commonly adopted during monitoring and routine care in neonatal intensive care units (NICUs))\cite{comparison2,finetuning2,finetuning3,enforcing}. This requires the creation of manually annotated infants' pose datasets. Annotation, however, remains a fundamental yet highly labor-intensive and time-consuming component of fully-supervised workflows which may be prone to errors \cite{annotation_pose_quality}. In PE, human annotators provide ground-truth labels and frequently encounter ambiguous cases caused by occlusions, poor lighting, or motion blur, all of which can compromise annotation quality and, consequently, model reliability. This observation is supported by preliminary studies reporting annotation noise in widely used human PE datasets, highlighting the detrimental impact of noisy labels (i.e., keypoints annotated at incorrect anatomical locations) on PE model performance \cite{influencefaultylabels}. In generic PE applications, such annotation errors typically lead only to modest reductions in aggregate localization metrics (e.g., Average Precision (AP) or Percentage of Correct Keypoints (PCK)). In the infant movement-assessment setting that motivates this work, however, their impact is fundamentally different. A mislocalized keypoint introduces an artificial displacement into the reconstructed trajectory of the corresponding joint. Because clinical motion biomarkers are computed directly from these trajectories, such artifacts may impact the very signal that the system is intended to analyze \cite{migliorelli_gma}. Consequently, ensuring that annotations are not corrupted becomes a critical prerequisite for the downstream clinical validity of motion-based assessment pipelines.

At the same time, learning with noisy labels became an established research area in deep learning, giving rise to a broad range of approaches, including specialized loss functions, noise-aware optimization strategies, and sample-selection methods ~\cite{song2025survey}. Among the latter, many techniques are based on the small-loss hypothesis \cite{coteaching,small_loss_2}, which assumes that samples incurring lower training losses during the early stages of optimization are more likely to be correctly labeled. These methods exploit the temporal evolution of the training loss, referred to as training dynamics, to distinguish clean from noisy annotations.

In a previous study \cite{remind}, we showed that sample-selection strategies based on the mean squared error (MSE) regression loss, conventionally used to train PE models, can effectively identify noisy keypoints. Specifically, we extracted per-keypoint features from the evolution of the regression loss during training and partitioned the entire set of annotated keypoints into clean and noisy subsets via unsupervised clustering. The main limitation of that approach, however, is its relatively low precision. This stems from the fact that high and unstable regression loss can arise from two distinct sources: genuinely mislocalized (noisy) keypoints, and correctly annotated keypoints that are intrinsically hard to localize, for instance due to occlusion, motion blur, or ambiguous limb configurations. The resulting training dynamics of noisy and hard-to-learn keypoints become difficult to separate, causing a substantial fraction of correctly annotated but hard-to-learn keypoints to be misclassified as noisy and thereby limiting the reliability of the approach as a data-curation step prior to PE model training.


The present work builds directly on this line of research and addresses its main limitation by encapsulating the task noisy-keypoint detection not in a regression framework, but in a classification one. First, the majority of effective training-dynamics-based noisy-label detection methods are developed for classification tasks. Second, and more specifically, an incorrectly localized keypoint induces a structural inconsistency between its spatial coordinates and the local visual evidence sampled at that location. We exploit this property by training a hybrid convolutional-attention classifier to predict the anatomical class of a keypoint (shoulder, elbow, wrist, hip, knee, and ankle) from its spatial coordinates and corresponding local visual features. We then extract training dynamics from the resulting cross-entropy (CE) classification loss and use them to construct per-keypoint feature representations, which are subsequently partitioned into clean and noisy subsets using an unsupervised clustering algorithm. These premises leads to our first research question:
\begin{itemize}
\item \textbf{RQ1:} To what extent do per-keypoint CE-based training dynamics reflect label reliability in human pose estimation?
\end{itemize}
We validate the proposed methodology on NeoPose, a newly collected dataset of preterm infant videos acquired during routine clinical practice and currently among the largest datasets specifically focused on hospitalized preterm infants. Beyond assessing this advantage on the controlled, homogeneous NeoPose setting, it remains an open question whether it extends to substantially more heterogeneous PE benchmarks, such as COCO \cite{coco}, where increased visual and postural complexity may reduce the separability between clean and noisy dynamics, and where the practical utility of filtering as a data-curation step — measured not only by detection accuracy but by its downstream effect on pose estimation performance — has yet to be assessed. This leads to our second research question:
\begin{itemize}
\item \textbf{RQ2:} Can the proposed noisy-label detection method generalize to heterogeneous PE datasets?
\end{itemize}
The main contributions of this work are summarized as follows:

\begin{itemize}
    \item We propose a novel framework for detecting noisy keypoint annotations in human pose estimation by exploiting training dynamics derived from the CE loss. We introduce a hybrid convolutional-attention architecture that predicts the anatomical class and body side of a keypoint from its spatial coordinates and local visual features, and derive a per-keypoint feature representation from its training dynamics for the unsupervised clustering of clean and noisy annotations.
    
    \item We provide experimental validation under two realistic and clinically motivated noise scenarios (local positional perturbations and left-right semantic swaps) across increasing noise levels, demonstrating that CE-based training dynamics consistently outperform MSE-based training dynamics..
    
    \item We show that the proposed framework generalizes to the COCO dataset and show that filtering training annotations using the proposed CE-based detector improves the downstream localization performance of a standard pose estimation model evaluated on a clean test set.
\end{itemize}

The remainder of this paper is organized as follows. Section \ref{sec:relatedwork} reviews related work on learning with noisy labels. Section \ref{sec:method} presents the proposed methodology, including the keypoint classifier architecture and the clustering-based separation of clean and noisy keypoints. Section \ref{sec:experimentalprotocol} describes the experimental protocol, including the noise injection procedures, the NeoPose dataset, the training settings, and the comparison methods. Section  \ref{sec:resultsanddisc}  presents and discusses the experimental results, organized around two research questions concerning the effectiveness and generalization of the proposed approach. Section \ref{sec:conclusion} concludes the paper and outlines directions for future research.

\section{Related work}
\label{sec:relatedwork}

\subsection{Learning with noisy labels}
It is by now widely recognized that deep neural networks and noisy labels form an uncomfortable pairing: the excess representational capacity of overparameterized architectures ultimately enables them to memorize incorrect annotations, causing a marked deterioration in generalization performance on held-out data \cite{zhang2017_noisylabels}. Yet foundational research has demonstrated that this memorization process is not random \cite{memorization_effect}. In the early phases of optimization, networks exhibit a preference for extracting generalizable patterns from correctly labeled examples — a phenomenon commonly referred to as the memorization effect — before gradually fitting corrupted annotations as training continues. Leveraging this observation, the small-loss hypothesis \cite{coteaching} formalizes the intuition that samples incurring consistently low loss values during training are strong candidates for correct labeling, and has catalyzed a broad family of sample-selection methods that exploit either the loss distribution in early epochs or, more broadly, the full temporal evolution of the training loss, commonly referred to as training dynamics. Within this paradigm, several approaches have been proposed to extract discriminative features from loss trajectories. Co-teaching \cite{coteaching} consists in training two networks simultaneously, using each to select low-loss examples for the other in order to reduce the mutual influence of noisy samples. More recent methods exploit richer temporal signals: the First-time k-epoch Learning metric\cite{latestopping} estimates label reliability by measuring how many epochs a sample requires before being consistently classified correctly; ChronoSelect \cite{chronoselect} employs a multi-stage temporal memory and trajectory analysis framework to segment data into clean, ambiguous, and noisy subsets; and an LSTM-based detector \cite{lstm_noisylabels} processes raw training-dynamics sequences to automatically flag mislabeled instances without relying on hand-crafted features. Notably, these approaches have been developed almost exclusively for classification settings, and comparatively little work has addressed noisy-label learning under regression objectives \cite{noisy_labels_regression1,noisy_labels_regression2,noisy_labels_regression3}, where the loss landscape differs qualitatively and the small-loss hypothesis applies less directly. Restricting attention further to human pose estimation, the landscape is even more sparse. The work of Schwarz et al. ~\cite{influencefaultylabels} provides a systematic empirical analysis of how different categories of annotation noise affect the performance of PE models, defining an extended taxonomy of label errors and quantifying their impact on standard localization metrics. On the animal PE side, ScarceNet \cite{scarcenet}  integrates a threshold-driven, small-loss sample selection step as the initial stage of a semi-supervised pipeline designed to train on scarce and potentially noisy pseudo-labels. Most closely related to the present work is our previous study \cite{remind}, in which we showed that MSE regression loss-based training dynamics can effectively distinguish clean from noisy keypoints in the infant PE setting, by extracting per-keypoint features from the loss trajectory and partitioning the full annotation pool into clean and noisy subsets via unsupervised clustering. 

\begin{figure}[bpt!]
\centering
\includegraphics[width=1\textwidth]{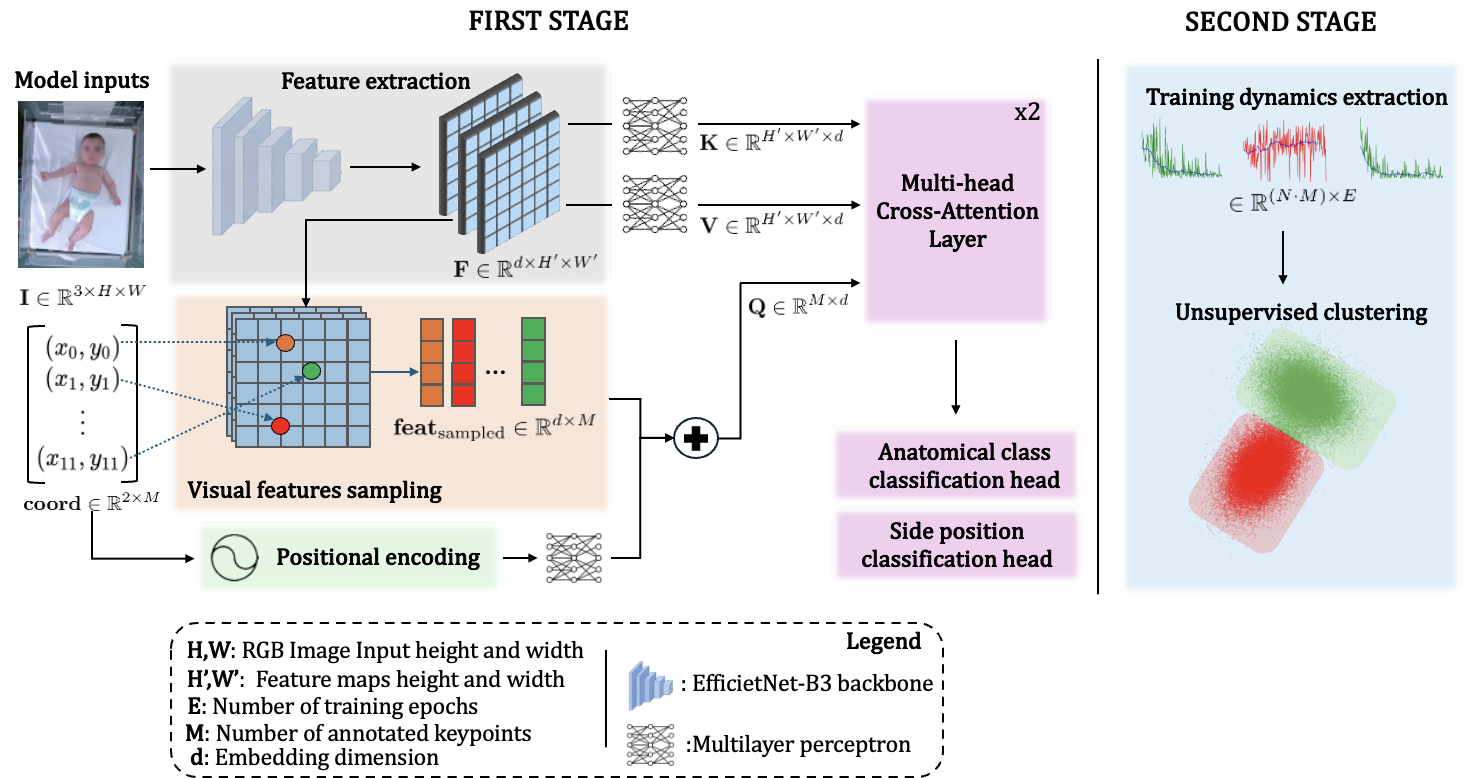}
\caption{Our proposed framework to detect noisy keypoints given a set of RGB images and their corresponding annotation set. It comprises two stages. In the first stage, a hybrid convolutional-attention classifier is trained to predict the anatomical category of a keypoint and its lateral position from the input image and its spatial coordinates. In the second stage, keypoint-wise features extracted from the trained model are used as input to an unsupervised clustering algorithm, which separates clean annotations from noisy ones.}
\label{fig:method_overview}
\end{figure}

\section{Method}
\label{sec:method}

Figure \ref{fig:method_overview} depicts an overview of our proposed methodology, aimed at detecting noisy keypoints in the context of infant PE. The proposed framework takes as input a set of images and their associated keypoint annotations, and detects which keypoints are noisy and which are clean. The framework comprises two stages, as shown in Fig. \ref{fig:method_overview}. In the first stage, a hybrid convolutional-attention classifier is trained to predict the anatomical category (shoulder, elbow, wrist, hip, knee, and ankle) of a keypoint from the input image and its spatial coordinates. In the second stage, keypoint-wise features extracted from the trained model are used as input to an unsupervised clustering algorithm, which separates clean annotations from noisy ones.

Let $D$ be a dataset of $N$ images, each containing a single infant lying in a supine position and annotated with a set of $M$ keypoints. Each keypoint is associated with 2D coordinates $\{(x_{m,n}, y_{m,n})\}_{m,n=1}^{M,N}$. We assume that a subset of these annotations across the entire dataset is noise-corrupted, meaning that the annotated 2D locations do not correspond to the true joint positions. We refer to these as noisy keypoints. Our approach builds on the sample selection paradigm and makes use of the well-established small-loss hypothesis. In our previous study \cite{remind}, we showed that MSE-based sample selection strategies can effectively distinguish between clean and noisy keypoints. Specifically, per-keypoint features are extracted from the training dynamics, and a clustering algorithm is applied to the entire pool of keypoints in the dataset to partition them into clean and noisy subsets. However, the main limitation of this approach is its relatively low precision. 

Motivated by the observation that training-dynamics-based sample-selection methods in the noisy-label learning literature have been developed for classification tasks, we propose a novel framework for noisy-keypoint detection.  The implementation of our proposed architecture is inspired by the work presented in \cite{keypoint_transformer}, where the authors introduce a method for estimating the 3D pose of two interacting hands from a single image, which decouples joint localization from joint identification. A key component of this framework—namely the module responsible for associating detected keypoints with their corresponding hand joints, which employs a self-attention mechanism operating on extracted CNN features—served as inspiration for the design of the proposed method. Specifically, we design a hybrid convolutional-attention keypoint classifier that, given a candidate keypoint location over the image and the image itself, predicts the anatomical class of the corresponding keypoint. Each keypoint is represented both by its spatial coordinates and by image features sampled at those locations. The CE loss is used to derive per-keypoint features, which are subsequently processed by a clustering algorithm. The key intuition is that incorrect keypoints correspond to erroneous spatial locations, which in turn yield misleading local visual features, creating a discrepancy between the input and its associated label that is reflected in the loss.

\subsection{First stage: Predicting anatomical class}
Given a single-infant RGB image $I \in \mathbb{R}^{H \times W \times 3}$ and a set of its $M$ keypoints with 2D coordinates $\{(x_m, y_m)\}_{m=1}^{M}$, the hybrid convolutional-attention keypoint classifier is trained to predict for each keypoint its anatomical class, represented by the label $j_m \in \{1, \dots, K\}$, and its lateral position (left/right), represented by the label $l_m \in \{0,1\}$. Note that $M$ represents the 12 keypoints (left/right shoulder, elbow, wrist, hip, knee, and ankle), retained from the  COCO standard 17-keypoints annotation format (5 facial keypoints extra) for each image. $K$ represents the 6 anatomical joint categories (shoulder, elbow, wrist, hip, knee, and ankle). Each keypoint of each image is then represented by an appearance and a spatial encoding. The appearance encoding is obtained by extracting local visual features from a pre-trained Efficient-Net model. More specifically, for each RGB image, the model backbone outputs feature maps  $\mathbf{F} \in \mathbb{R}^{C \times H' \times W'}$, where C is the number of feature channels, $H'$ the height of the feature maps and $W'$ the width of the feature maps, and  a 1x1 convolution projects the extracted feature map to a fixed embedding dimension $d$, obtaining $\mathbf{F'} \in \mathbb{R}^{d \times H' \times W'}$. The keypoint coordinates $\{(x_m, y_m)\}_{m=1}^{M}$ are mapped to the spatial resolution of the feature maps, which are sampled at these locations via bilinear interpolation. The output of the sampling operation is a feature vector $\mathbf{f}_k \in \mathbb{R}^{d}$ for each keypoint. Similarly, the coordinates of each keypoint are encoded using a sinusoidal positional encoding function $\gamma(\cdot)$, similar to \cite{keypoint_transformer}, and projected into the same embedding space through a 2-layer Multi-Layer Perceptron (MLP):
\begin{equation}
\mathbf{pos_k} = \mathrm{MLP}(\gamma(\mathbf{p}_k)).
\end{equation}

The final representation is obtained by combining spatial and visual information $\mathbf{q}_k = \mathbf{pos_k} + \mathbf{f}_k$. Stacking all keypoint representations, we obtain $\mathbf{Q'} = [\mathbf{q'}_1, \dots, \mathbf{q'}_M] \in \mathbb{R}^{M \times d}$.
A subsequent cross-attention module treats these keypoints encodings as queries that can gather information from the entire feature map. The keypoints are linearly projected to form the query vectors:
\begin{equation}
\mathbf{Q} = \mathbf{Q'} \mathbf{W}_q, \quad \mathbf{W}_q \in \mathbb{R}^{d \times d}.
\end{equation}
The feature map is flattened into $\mathbf{F}_{\text{flat}} \in \mathbb{R}^{(H'W') \times d}$ and projected into key and value representations:
\begin{equation}
\mathbf{K} = \mathbf{F}_{\text{flat}} \mathbf{W}_k, \quad \mathbf{V} = \mathbf{F}_{\text{flat}} \mathbf{W}_v,
\end{equation}
where $\mathbf{W}_k \in \mathbb{R}^{d \times d}$, $\mathbf{W}_v \in \mathbb{R}^{d \times d}$, $\mathbf{K}, \mathbf{V} \in \mathbb{R}^{(H'W') \times d}$. \\
Cross-attention is computed as:
\begin{equation}
\text{Attention}(Q,K,V) = \text{softmax}\left(\frac{\mathbf{Q}\mathbf{K}^T}{\sqrt{d}}\right)\mathbf{V}
\end{equation}
Similar to \cite{attention_allyouneed}, the cross-attention features are added to $\mathbf{Q}$ to create a residual connection. The resulting context-aware features are used to retrieve the anatomical category and side position $(j_m, l_m)$ of each of the $M$ keypoint. To this end, we employ two separate linear classification heads, the first predicting $j_m$ , while the second predicting $l_m$. Both heads are trained using CE loss. For a single image, the total loss is defined as a weighted sum of the two terms:
\begin{equation}
\mathcal{L} = \mathcal{L}_{\text{joint}} + \lambda \, \mathcal{L}_{\text{side}},
\end{equation}

where
\begin{equation}
\mathcal{L}_{\text{joint}} = \frac{1}{M} \sum_{m=1}^{M} \mathrm{CE}\big(\hat{j}_m, j_m\big),
\end{equation}

\begin{equation}
\mathcal{L}_{\text{side}} = \frac{1}{M} \sum_{m=1}^{M} \mathrm{CE}\big(\hat{l}_m, l_m\big),
\end{equation}

where $\hat{j}_m$ and $\hat{l}_m$ denote the predicted logits for anatomical class and side position, respectively, and $\lambda$ controls the contribution of the latter. From now on, for simplicity, we will refer to the classification loss on lateral positions as the side loss.

\subsection{Second stage: Noisy and clean keypoints clustering}
\label{subsec:noisycleansub}

As the training progresses, for each keypoint $m$ in each image $n$, the joint loss and side loss values are recorded at every epoch, forming two loss trajectory vectors, $\mathbf{L}^{\text{joint}}_{m,n} \in \mathbb{R}^{E}$ and $\mathbf{L}^{\text{side}}_{m,n} \in \mathbb{R}^{E}$, respectively, where $E$ is the number of training epochs. To reduce stochastic fluctuations, both trajectories are smoothed using a moving average filter. Based on experimental observations presented in \cite{remind}, clean and noisy keypoints typically exhibit different learning trajectories. Thus, we make use of a set of features derived from the loss trajectories for each keypoint at the end of training.
The first feature, $\Delta l$, measures the overall learning dynamics by quantifying the normalized peak-to-trough loss variation:
\begin{equation}
\Delta l_{m,n} =
\frac{\max(\mathbf{L}_{m,n}) - \min(\mathbf{L}_{m,n})}
{\max(\mathbf{L}_{m,n}) + \min(\mathbf{L}_{m,n})}.
\end{equation}
The second feature, $\Delta t$, captures temporal behavior by measuring when the minimum loss is reached relative to the maximum:
\begin{equation}
\Delta t_{m,n} =
\frac{\operatorname{argmin}_e \mathbf{L}_{m,n} - \operatorname{argmax}_e \mathbf{L}_{m,n}}{E},
\end{equation}

Both features are extracted from the joint loss and the side loss. They will be referred to as $(\Delta l_{side}, \Delta t_{side},\Delta l_{joint}, \Delta t_{joint})$.

Given these features, we reshape our dataset into $(N \times M) \times 4$, where $N$ denotes the number of images and $M$ the number of keypoints per image. Each keypoint is represented in a 4-dimensional feature space defined by $(\Delta l_{side}, \Delta t_{side},\Delta l_{joint}, \Delta t_{joint})$. We then apply a vanilla K-means clustering to partition the keypoints in this feature space. Since, empirically, the feature values tend to decrease as annotation noise increases, the cluster characterised by the lower centroid is identified as containing noisy keypoints.

\section{Experimental Protocol}
\label{sec:experimentalprotocol}
\subsection{Noise injection}

The work presented in \cite{influencefaultylabels} defines an extended error taxonomy of label noise in PE. Setting aside noise arising from annotated occluded keypoints and unannotated visible keypoints, in our experiments we consider two plausible noise scenarios: incorrect position and left–right swap. The first noise scenario is designed to reflect typical human inaccuracies during manual annotation, such as small shifts caused by mouse or touchpad imprecision. Following the work of Johnson et al. \cite{noisy_pose_estimation2011}, we model keypoint localization errors as an isotropic Gaussian distribution of vertical and horizontal displacements, i.e., by adding zero-mean Gaussian noise to the ground-truth joint locations, with variance ranging from 0.15\% to 0.25\% of the infant bounding box size. We refer to this as the \textbf{random-noise} scenario. The second noise scenario models structural confusion between the left and right parts of the body. This reflects the common ambiguity regarding whether “left” and “right” refer to the subject in the image or the annotator’s perspective. To simulate this type of noise, we randomly select a subset of samples from $D$ and swap left keypoints with right ones. We refer to this as the \textbf{swapping-noise} scenario. For both scenarios, we evaluate multiple configurations with progressively increasing levels of noise (from 10\% to 40\%), a range selected based on the most commonly used noise levels in the noisy-label classification literature (typically 20\% and 40\% for both instance-dependent and symmetric noise settings).

\subsection{Dataset}

In this study, we collect and curate a dataset of 65 RGB video recordings of 65 infants born preterm, who were recorded in a state of active wakefulness at term equivalent age or right before discharge in the neonatology department of the G. Salesi Hospital in Ancona, Italy. The 65 preterm infants in this dataset represent a comparatively robust sample, given how resource-intensive and logistically challenging clinical data collection is in this population. The dataset shows high variability in terms of gestational age ($31.87 \pm 3.77$), weight ($2 \pm 0.79$ kg), and length ($44.13 \pm 4.12$ cm). The infants were identified by clinicians among those in stable condition and capable of spontaneous movement. Videos are recorded using an Astra Mini S-Orbbec RGB-D camera, positioned approximately $\sim$50 cm above the crib, with a spatial resolution of 640×480 pixels and a frame rate of 30 fps. Each video refers to a single infant, recorded shortly before hospital discharge. On average, approximately 116 frames per infant were randomly sampled from each continuous five-minute recording in order to maximize pose diversity, totalling 7563 RGB frames.  Manual annotation was carried out by a team of experts neonatologists following the standard COCO keypoint annotation format.
\subsection{Training settings}

We use the backbone of EfficientNet-B3 as feature extractor, initialized with ImageNet-pretrained weights. The model is optimized using the Adam optimizer for 40 epochs. The backbone is fine-tuned using a reduced learning rate of $4 \times 10^{-5}$, while all task-specific layers are optimized with a higher learning rate of $2 \times 10^{-4}$. Weight decay is set to $1 \times 10^{-5}$ for all parameters. The learning rate schedule consists of a 10-epoch linear warm-up, followed by cosine decay down to a minimum learning rate of $1 \times 10^{-5}$. Keypoint representations are learned in a 128-dimensional embedding space. Input coordinates are enriched using a sinusoidal positional encoding with 8 frequency bands and projected via an MLP into the shared embedding space. The attention module consists of a 2-layer cross-attention stack, with 2 attention heads per layer, enabling interaction between keypoint embeddings and global image features extracted from the backbone. The weight $lambda$ of the left/right classification loss is gradually increased during training up to 0.5, allowing the model to first focus on learning robust keypoint representations before incorporating directional information. Inputs are resized to $300 \times 300$, and training is performed with a batch size of 16. Experiments are conducted using Python 3.11 and PyTorch 2.7.1 on a single NVIDIA Ampere A100 GPU (64 GB).

\subsection{Comparisons}
\label{sec:comparisons}

We structure the experimental comparisons in two steps. First, we identify the most discriminative subset of CE-based per-keypoint features for the clustering stage. Second, using the best-performing feature configuration, we compare the proposed CE-based framework against our previously proposed MSE-based method on two datasets differing in visual complexity and population homogeneity: NeoPose and COCO.

\subsubsection{Feature selection for noisy-keypoint clustering}
\label{sec:feature_selection}

The clustering stage separates clean from noisy keypoints using a four-dimensional descriptor set ($\Delta l_{side}$, $\Delta t_{side}$, $\Delta l_{joint}$, $\Delta t_{joint}$) derived from the CE loss trajectories. Since not all descriptors are equally informative for every noise type, we ablate representative feature subsets and evaluate detection F1-score separately for the random-noise and swapping-noise scenarios on NeoPose, across all noise levels. This analysis identifies the most discriminative descriptors for each noise type and a single, noise-agnostic configuration that performs reliably under both, without requiring prior knowledge of the corruption type affecting a given dataset. This configuration is then adopted unchanged for all subsequent experiments.

\subsubsection{Comparison with the MSE-based method and cross-dataset evaluation on COCO}
\label{sec:mse_comparison}

We compare the CE-based framework against our previously proposed MSE-based method, which follows the same sample-selection paradigm but derives training dynamics from the MSE regression loss of an  HRNet model rather than from a classification loss. HRNet was selected as a widely adopted regression-based PE architecture, also commonly used in the infant PE literature \cite{cacciatore}, and was trained from scratch using the MMPose toolbox under its default protocol. The resulting ($\Delta l$, $\Delta t$) descriptors are partitioned using the same K-means clustering procedure, isolating the effect of the loss type from the clustering stage itself. On NeoPose, both methods are evaluated under random and swapping noise across all injected noise levels. On COCO, given its higher heterogeneity, we restrict the comparison to the random-noise scenario, assessing both detection performance and the downstream effect of removing detected noisy keypoints on the localization accuracy of an HRNet model evaluated on a clean test set.

\subsection{Performance metrics}
\label{sec:metrics}

For the evaluation metrics we use the $F1-score$ to assess the quality of noisy-label detection. As the harmonic mean of $Precision$ and $Recall$, it provides a single measure that balances the impact of false positive (i.e., correctly annotated keypoints incorrectly classified as noisy) and false negatives (i.e., noisy keypoints incorrectly classified as clean). 

To quantify the separability of clean and noisy loss trajectories during training, we introduce the Separability Metric (SeM) At the end of training, we compute the mean loss trajectory for both groups, along with their corresponding epoch-wise standard deviations. For each epoch $t$, we model the loss distributions of the clean and noisy keypoint groups as Gaussians $\mathcal{N}(\mu_c(t), \sigma_c(t)^2)$ and $\mathcal{N}(\mu_n(t), \sigma_n(t)^2)$, where $\mu$ and $\sigma$ are the empirical mean and standard deviation of the loss values within each group. The overlap area between the two distributions is computed and averaged across epochs, yielding a single scalar score. Lower SeM values indicate better separation between clean and noisy keypoint dynamics.

Downstream pose estimation performance, as a measure of the benefit of noisy-label filtering, is evaluated using a standard PE benchmark metric, the COCO Average Precision ($AP$). The COCO $AP$ is calculated as the area under the precision-recall curve, averaged over multiple Object Keypoint Similarity ($OKS$) thresholds~\cite{maji2022yolo}, ranging from 0.50 to 0.95 in increments of 0.05.

\section{Results and discussion}
\label{sec:resultsanddisc}

We organize the experimental findings into three parts. We first report the outcome of the feature-selection analysis, which determines the configuration used throughout the remainder of the evaluation. We then address our two research questions: \textbf{RQ1}, concerning the effectiveness of CE-based training dynamics for noisy-label detection on the target infant population, and \textbf{RQ2}, concerning the generalization of the proposed method to a heterogeneous PE benchmark.

\begin{table}[h!]
\centering
\caption{Feature selection for noisy-keypoint clustering. F1-scores are reported separately for the random-noise and swapping-noise scenarios, averaged across the four injected noise levels (10\%, 20\%, 30\%, 40\%).}
\renewcommand{\arraystretch}{1.2}
\begin{tabular}{c|c|c|c|cc}
\hline
\textbf{$\Delta t_{side}$} &
\textbf{$\Delta l_{side}$} &
\textbf{$\Delta t_{joint}$} &
\textbf{$\Delta l_{joint}$} &
\multicolumn{2}{c}{\textbf{$F1-score$}} \\ \cline{5-6}
 &  &  &  & \textbf{Randomic noise} & \textbf{Swapping noise} \\ \hline

\checkmark &  &  &  &
$0.749 \pm 0.012$ &
$0.596 \pm 0.088$ \\ 

 & \checkmark &  &  &
$0.521 \pm 0.064$ &
$0.779 \pm 0.104$ \\ 

 &  & \checkmark &  &
$0.896 \pm 0.032$ &
$0.001 \pm 0.001$ \\ 

 &  &  & \checkmark &
$0.491 \pm 0.054$ &
$0.200 \pm 0.063$ \\ 

\checkmark & \checkmark &  &  &
$0.757 \pm 0.044$ &
$0.812 \pm 0.081$ \\ 

 &  & \checkmark & \checkmark &
$0.840 \pm 0.016$ &
$0.135 \pm 0.085$ \\ 

\checkmark & \checkmark & \checkmark & \checkmark &
$\textbf{0.908} \pm 0.003$ &
$\textbf{0.812} \pm 0.081$ \\ 

\end{tabular}
\label{tab:ablation_side_joint}
\end{table}




\begin{table}[h!]
\centering
\caption{$F1$-score on the NeoPose dataset under increasing levels of random and swapping noise (MSE- vs CE-based methods).}
\renewcommand{\arraystretch}{1.2}
\begin{tabular}{c|cc|cc}
\hline
\textbf{Noise (\%)} & \multicolumn{2}{c|}{\textbf{Randomic Noise}} & \multicolumn{2}{c}{\textbf{Swapping Noise}} \\ \cline{2-5}
 & \textbf{MSE-based} & \textbf{CE-based} & \textbf{MSE-based} & \textbf{CE-based} \\ \hline

10 & 0.558 & \textbf{0.906} & 0.510 & \textbf{0.894} \\ \hline
20 & 0.654 & \textbf{0.909} & 0.861 & \textbf{0.885} \\ \hline
30 & 0.672 & \textbf{0.911} & 0.743 & \textbf{0.771} \\ \hline
40 & 0.674 & \textbf{0.904} & 0.632 & \textbf{0.699} \\ 

\end{tabular}
\label{tab:f1_score_infant}
\end{table}

\subsection{Feature selection for noisy-keypoint clustering}
Table \ref{tab:ablation_side_joint} reports the mean $F1-score$, averaged across all injected noise levels, obtained by K-means clustering on each candidate feature subset, separately for the random-noise and swapping-noise scenarios.

In the random-noise scenario, the most discriminative single feature is $\Delta t_{joint}$ ($0.896 \pm 0.032$), consistent with the intuition that a coordinate shift primarily disrupts the local visual evidence available for anatomical-class prediction. Side-loss features, by contrast, are only moderately informative in isolation ($\Delta t_{side}$: $0.749 \pm 0.012$; $\Delta l_{side}$: $0.521 \pm 0.064$), which is expected since random positional perturbations do not alter the left/right assignment of a keypoint, leaving the side-loss signal largely uninformative with respect to this noise type. The full feature set achieves the highest performance ($0.908 \pm 0.003$), indicating that side-loss features, while non-informative in isolation, do not degrade the clustering when included alongside the more discriminative joint features.

In the swapping-noise scenario, the pattern is complementary. $\Delta t_{joint}$ alone collapses to near-zero $F1-score$ ($0.001 \pm 0.001$), and the joint feature pair performs poorly ($0.135 \pm 0.085$), confirming that a left/right swap leaves the anatomical category of a keypoint entirely unchanged, rendering joint-classification dynamics blind to this type of corruption. Conversely, $\Delta l_{side}$ emerges as the most discriminative single descriptor ($0.779 \pm 0.104$), and combining both side-loss features yields the best single-group result ($0.812 \pm 0.081$), a performance that is fully preserved when joint features are added to form the complete set ($0.812 \pm 0.081$). This confirms that the clustering mechanism treats the non-informative joint features as irrelevant signal without penalizing overall detection quality.

Taken together, these results reveal a clear functional specialization: joint-loss temporal dynamics is the primary signal for detecting random positional corruption, while side-loss dynamics are indispensable for detecting semantic left/right swaps. 

The four-dimensional feature set $(\Delta t_{side}, \Delta l_{side}, \Delta t_{joint}, \Delta l_{joint})$ achieves the best or joint-best performance under both noise types, without requiring prior knowledge of which corruption mechanism affects a given dataset. This noise-agnostic behavior is practically relevant, since in real annotation pipelines the type and proportion of label corruption are typically unknown a priori. We therefore adopt this complete feature configuration for all subsequent experiments.

\begin{figure}[h!]
\centering
\includegraphics[width=1\textwidth]{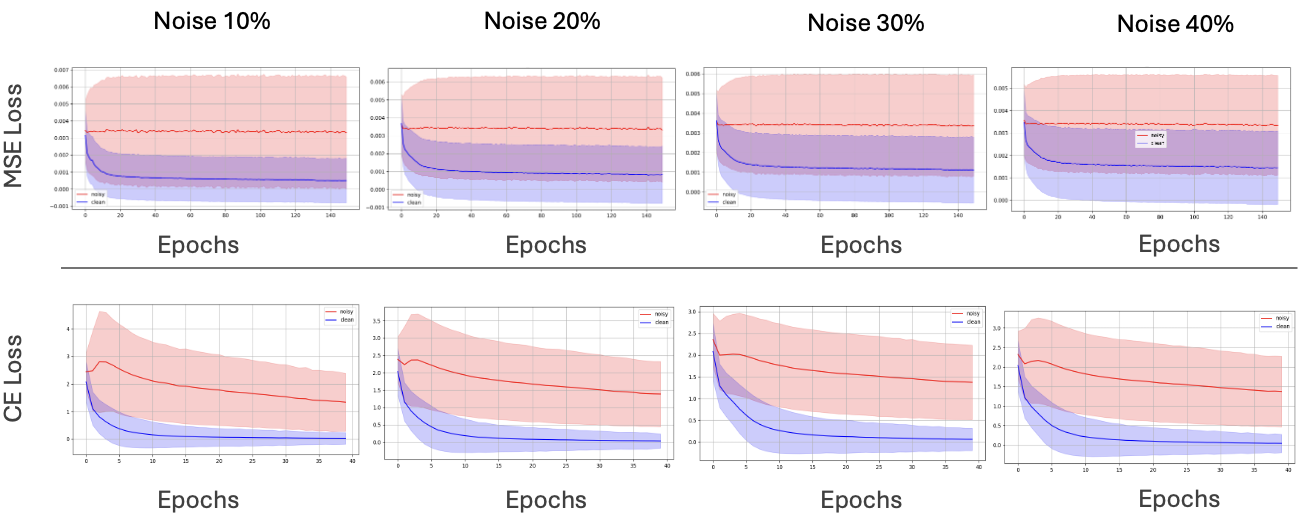}
\caption{Mean loss trajectories for clean (blue) and noisy (red) keypoints under the random-noise scenario for NeoPose dataset, reported with their epoch-wise standard deviation across training. Each column corresponds to an increasing noise level (10\%, 20\%, 30\%, 40\%). The first row shows the MSE regression loss trajectories extracted from HRNet, used in the MSE-based baseline. The second row shows the CE classification loss trajectories extracted from the proposed keypoint classifier.}
\label{fig:visual_resuls_loss}
\end{figure}

\begin{table}[h!]
\centering
\caption{Separability metric (SeM) between clean and noisy keypoint loss trajectories under the random-noise scenario, for increasing noise levels (10\%, 20\%, 30\%, 40\%). The first row reports values obtained from the MSE regression loss trajectories of the MSE-based baseline; the second row from the CE classification loss trajectories of the proposed method.Lower values indicate better separation between clean and noisy dynamics.}
\begin{tabular}{lcccc}
\hline
 & Noise 10\% & Noise 20\% & Noise 30\% & Noise 40\% \\ \hline
MSE-based & 0.4524 & 0.5385 & 0.5843 & 0.6199 \\ \hline
CE-based & \textbf{0.2413} & \textbf{0.2475} & \textbf{0.2615} & \textbf{0.2770} \\ \hline
\end{tabular}
\label{tab:separability}
\end{table}


\begin{figure}[h!]
\centering
\includegraphics[width=1\textwidth]{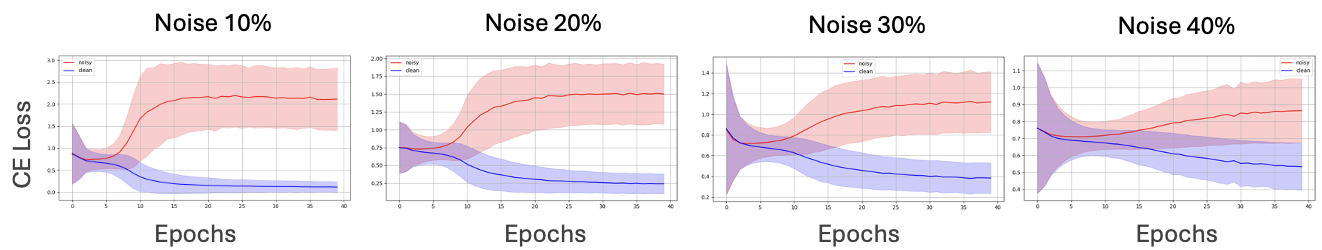}
\caption{Mean side loss trajectories for clean (blue) and noisy (red) keypoints under the swapping-noise scenario, reported with their epoch-wise standard deviation across training. Each column corresponds to an increasing noise level (10\%, 20\%, 30\%, 40\%). Trajectories are extracted from the side classification loss of the proposed CE-based method.}
\label{fig:side_loss}
\end{figure}

\subsection{RQ1: To what extent do per-keypoint CE-based training dynamics reflect label reliability in human PE?}

Table~\ref{tab:f1_score_infant} reports the $F1-score$s obtained by the MSE-based and CE-based detection pipelines across increasing levels of random and swapping noise on NeoPose. Across all configurations of the random-noise and swapping-noise scenario the CE-based approach substantially and consistently outperforms the MSE-based baseline.
This behavior may be due to the logarithmic penalization of misclassifications induced by the cross-entropy loss, in contrast to the quadratic penalization imposed by MSE~\cite{ce_vs_mse_loss}. In particular, MSE may suffer from gradient attenuation for small perturbations around the true keypoint location, resulting in less discriminative training dynamics.  This is qualitatively confirmed in Fig.\ref{fig:visual_resuls_loss}, where the CE-based loss trajectories (second row) display markedly wider separation between the mean clean and noisy curves than their MSE-based counterparts (first row), and in Table \ref{tab:separability}, where the proposed separability metric SeM is consistently lower (i.e., better separated) for the CE-based dynamics across randomic -noise levels.
A secondary observation concerns the comparatively high and stable performance observed under random noise even at elevated corruption levels. We interpret this as a consequence of the structural properties of NeoPose: because frames are sampled from continuous video recordings of infants lying in a constrained supine position, the dataset exhibits limited pose variability and strong spatial regularities among keypoints. As a result, small, localized coordinate perturbations —- the noise model adopted for the random-noise scenario -- tend to remain within the same semantic neighborhood as the true keypoint, only marginally disrupting the visual-spatial consistency exploited by the classifier. This property does not extend to the swapping-noise scenario, since exchanging left and right labels alters the semantic structure of the supervision signal rather than introducing a local geometric perturbation. As the proportion of swapped keypoints grows, the model is exposed to increasingly conflicting supervision for symmetric body parts, and the problem becomes progressively more degenerate as left/right symmetry is approached (Fig. \ref{fig:side_loss}), explaining the steeper performance decay observed for swapping noise at higher corruption levels.

\begin{table}[tbp!]
\centering
\caption{$F1-score$ on COCO pose estimation under increasing levels of randomic noise}
\renewcommand{\arraystretch}{1.2}
\begin{tabular}{c|cc}
\hline
\textbf{Noise (\%)} & \multicolumn{2}{c}{\textbf{Random Noise}} \\ \cline{2-3}
 & \textbf{MSE Based} & \textbf{CE Based} \\ \hline

10 & 0.387 & 0.510 \\ \hline
20 & 0.560 & 0.670 \\ \hline
30 & 0.628 & 0.733 \\ \hline
40 & 0.661 & 0.784 \\ 

\end{tabular}
\label{tab:coco_table_f1}
\end{table}

\begin{figure}[h!]
\centering
\includegraphics[width=1\textwidth]{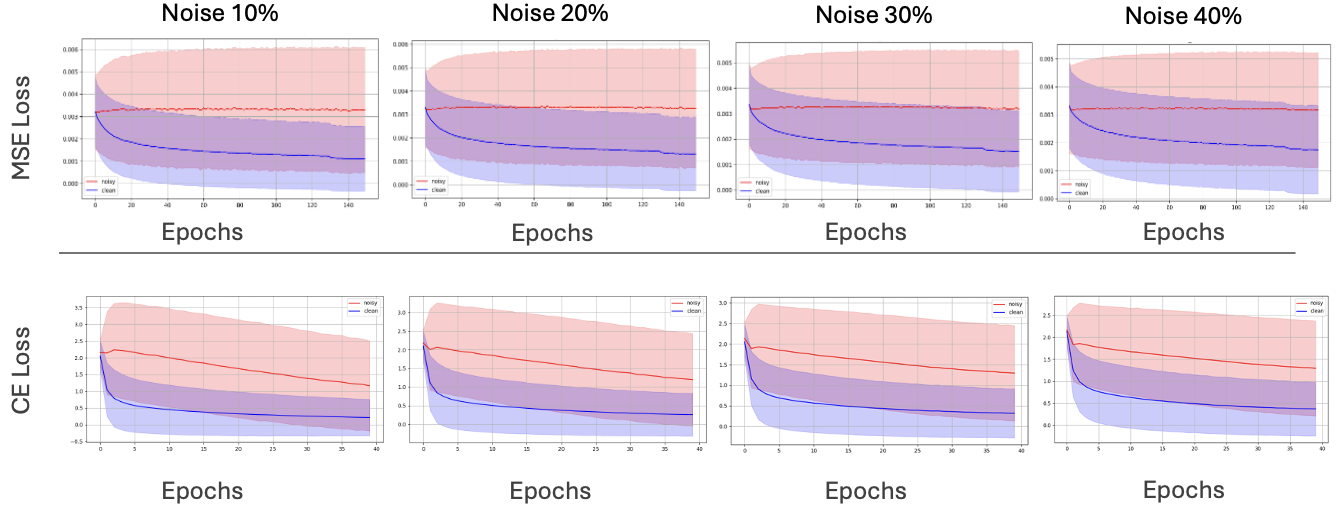}
\caption{Mean loss trajectories for clean (blue) and noisy (red) keypoints under the random-noise scenario for COCO dataset, reported with their epoch-wise standard deviation across training. Each column corresponds to an increasing noise level (10\%, 20\%, 30\%, 40\%). The first row shows the MSE regression loss trajectories extracted from HRNet, used in the MSE-based baseline. The second row shows the CE classification loss trajectories extracted from the proposed keypoint classifier.}
\label{fig:coco_mse_ce}
\end{figure}

\begin{figure}[h!]
\centering
\includegraphics[width=0.85\textwidth]{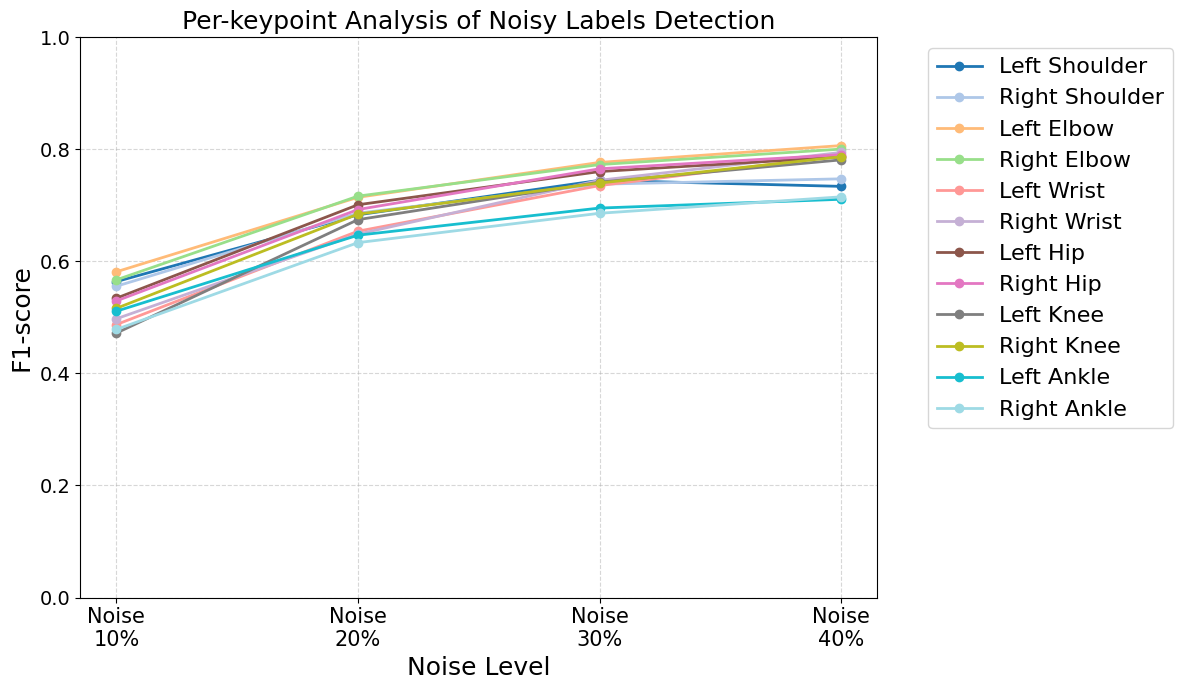}
\caption{Per-keypoint noisy-label detection F1-score on the COCO dataset under the random-noise scenario, reported for each of the 12 annotated keypoints across the four noise levels (10\%, 20\%, 30\%, 40\%).}
\label{fig:kp_analysis}
\end{figure}

\subsection{RQ2: Can the proposed noisy-label detection method generalize to heterogeneous PE datasets?}

To assess generalization beyond the relatively homogeneous and controlled NeoPose setting, we repeat the random-noise evaluation on a 14K-image subset of COCO \cite{coco}(Table \ref{tab:coco_table_f1}). As expected, absolute detection performance decreases relative to NeoPose for both methods, which we attribute to the substantially higher visual and postural heterogeneity of COCO  (i.e., multiple subjects, varied viewpoints, clothing, occlusion patterns, and backgrounds) compared to the structurally constrained infant-in-crib scenario. This is corroborated qualitatively by Fig. \ref{fig:coco_mse_ce}, which shows visibly reduced separability between the mean loss trajectories of clean and noisy keypoints relative to NeoPose (\ref{fig:visual_resuls_loss}).
Despite this expected drop, three results support the robustness and value of the proposed approach. First, the CE-based pipeline consistently and substantially outperforms the MSE-based baseline across all tested noise levels, confirming on an independent, heterogeneous benchmark the same trend observed on NeoPose, and reinforcing the generality of the underlying hypothesis that classification-based training dynamics are more sensitive to keypoint-level label corruption than regression-based ones. Second, we observe a counter-intuitive trend whereby detection performance improves as the noise level increases, rather than degrading as one might expect. We hypothesize this reflects a class-balance effect intrinsic to unsupervised K-means clustering: because the algorithm receives no prior information about the relative size of the clean and noisy partitions, it is implicitly biased toward producing clusters of comparable size. At low noise levels, where clean keypoints vastly outnumber noisy ones, this bias works against correct partitioning; as the noise level increases and the two classes become more balanced, the clustering more closely matches the true underlying partition, improving F1-score. This suggests that future work could benefit from clustering or thresholding strategies that explicitly account for, or are invariant to, class imbalance.
Third, a per-keypoint breakdown (Fig. \ref{fig:kp_analysis}) reveals statistically significant heterogeneity in detection difficulty across anatomical landmarks (Friedman test, p<0.05). Post-hoc pairwise comparisons with FDR correction identify the left and right ankle as significantly harder to classify as clean/noisy than all other keypoints. This is consistent with established findings in the PE literature, which report that distal joints are disproportionately affected by localization ambiguity, motion blur, and occlusion \cite{distal_joints1}.


\subsubsection{Downstream impact of noisy-label filtering on PE performance}

The results presented so far quantify the ability of the proposed CE-based framework to detect noisy keypoints, but do not directly establish whether removing these detected annotations is actually beneficial for training a downstream pose estimation model. To address this practical question, we train HRNet — the same regression-based PE architecture used to derive the MSE-based training dynamics in Section 4.4 — under each of the four COCO noise configurations (10\%, 20\%, 30\%, 40\% random noise), both (i) on the full, noise-corrupted training set, and (ii) on the corresponding training set after removing the keypoints flagged as noisy by the CE-based detection pipeline. Both models are evaluated on a held-out, clean COCO test set, using COCO AP. The results reveal a noise-level-dependent effect of filtering. At the lowest noise level (10\%), training on the full, unfiltered training set yields a higher COCO AP (0.633) than training on the CE-filtered subset (0.601). Conversely, at 20\%, 30\%, and 40\% noise, filtering consistently improves downstream performance relative to training on the unfiltered data, raising COCO AP from 0.544 to 0.618 at 20\% noise, from 0.542 to 0.599 at 30\% noise, and from 0.522 to 0.573 at 40\% noise. This pattern is consistent with, and reinforces, the detection results already discussed: at 10\% noise, the CE-based detector achieves its lowest noisy-label detection $F1-score$ among the four configurations (Table 3), reflecting the class-imbalance bias of K-means discussed earlier — at low noise proportions, the clustering algorithm tends to produce more balanced (and therefore less precise) clusters, leading to a higher rate of false positives, i.e., correctly annotated keypoints erroneously discarded as noisy. In this regime, the loss of legitimately clean training signal outweighs the benefit of removing the comparatively few truly noisy keypoints, resulting in a net degradation of downstream performance. As the noise level increases to 20–40\%, detection $F1-score$ improves accordingly (Table \ref{tab:coco_table_f1}), the clustering more closely approximates the true noisy/clean partition, and the benefit of removing an increasingly large and increasingly correctly identified set of corrupted annotations comes to dominate, translating into a measurable improvement in downstream PE accuracy. These findings provide an important practical caveat: noisy-label detection performance, on its own, is not a sufficient indicator of training utility — its downstream benefit also depends on the trade-off between detection precision and the proportion of valid training signal that any filtering procedure inevitably removes. They further suggest that the proposed framework is most beneficial as a data-curation step when annotation noise is moderate-to-high ($\geq$ 20\% in our setting), while at very low noise levels a more conservative, precision-oriented filtering strategy may be preferable to avoid discarding informative clean annotations.

\section{Conclusion}
\label{sec:conclusion}
Accurate infant PE models require large-scale supervised training, and training in turn requires manually annotated keypoints. Annotation, however, is an inherently error-prone process, and this limitation becomes especially critical in our clinical application: a noisy keypoint can distort the joint trajectories from which motion biomarkers are computed, undermining the reliability of the resulting clinical assessment. Motivated by this problem, in this work we introduce a novel framework for detecting noisy keypoint annotations, reformulating the task as an anatomical classification problem and exploiting cross-entropy training dynamics through a hybrid convolutional-attention architecture. This approach substantially improves upon a previously proposed MSE-based method, achieving higher F1-score in noisy-keypoints detection across both random-positional and left–right swap noise scenarios on NeoPose, and generalizing to the more heterogeneous COCO benchmark, where filtering detected noisy keypoints also translates into measurable gains in downstream pose estimation accuracy. Despite these promising results, some limitations remain. The K-means clustering stage exhibits a class-balance bias that penalizes detection precision at low noise levels, and the noise scenarios considered, while clinically motivated, do not cover all sources of annotation error encountered in practice, such as those that can arise from occlusions or ambiguous poses. Future work will explore clustering strategies that are explicitly robust to class imbalance, extend the noise taxonomy to additional real-world annotation error types, and investigate integrating the proposed detection framework as an online data-curation step within the PE training loop, in order to improve the reliability of infant PE models used to support clinical motion assessment.

\bibliographystyle{elsarticle-num}
\bibliography{biblio}

\end{document}